# Analyzing User Activities Using Vector Space Model in Online Social Networks

Dhrubasish Sarkar[1], Premananda Jana[2]

[1]School of Management & Allied Courses, Amity University Kolkata
[2]Department of CSE, MCKV Institute of Engineering, Liluah, Howrah
{dhrubasish@inbox.com[1], prema_jana@yahoo.com[2]}

[1]Corresponding Author. Mobile No. +919830571808.

**Abstract**

The increasing popularity of internet, wireless technologies and mobile devices has led to the birth of mass connectivity and online interaction through Online Social Networks (OSNs) and similar environments. OSN reflects a social structure consist of a set of individuals and different types of ties like connections, relationships, interactions etc among them and helps its users to connect with their friends and common interest groups, share views and to pass information. Now days the users choose OSN sites as a most preferred place for sharing their updates, different views, posting photographs and would like to make it available for others for viewing, rating and making comments. The current paper aims to explore and analyze the association between the objects (like photographs, posts etc) and its viewers (friends, acquaintances etc) for a given user and to find activity relationship among them by using the TF-IDF scheme of Vector Space Model. After vectorization the vector data has been presented through a weighted graph with various properties.

**Keywords:** vector space model, term-frequency inverse document frequency, online social network, graph visualization, data mining

## I. Introduction

Now days, with the immense growth and popularity of internet, wireless technologies and mobile devices, the Online Social Networks (OSNs) and similar applications have rapidly promoted networked interaction environment where people can interact with each other easily and instantly. An OSN reflects a social structure consists of a set of individuals and different types of association or ties among them like connections, interactions, relationship etc [1]. Some of the OSN sites are popular in some countries while some of them are having global reach. Some of the sites are designed to cater the need of a specific interest group while others are general in nature [2]. Generally the OSN sites enable its uses to post their various updates or comments, upload various documents, pictures etc which are available to their friends and acquaintances for viewing, sharing and making comments, express liking etc. Those data mostly are non tabular in nature. These types of unstructured data are first converted to tabular data and then various data mining algorithms can be used to analyze the relationships among them. To convert such non tabular, text data into tabular format, the vectorization process can be used. The proposed model utilizes one such well-known vectorization known as Vector Space Model (VSM) introduced by Salton, Wong and Yang [3]. In this paper, the term frequency - inverse document frequency (TF-IDF) weighting scheme has been used on a set of collected data on user activities as friends or acquaintances and then the processed, tabular data has been presented through a weighted graph using Gephi [8] software tool. The paper is organized as follows. Section II discusses the concepts of Vector Space Model and TF-IDF scheme. Section III discusses Related Works. Section IV discusses the Data Models and experimental results. Last section contains conclusions and future scope.

## II. Basic Concepts of Vector Space Model and TF-IDF

The Vector Space Model is a standard and well known method for vectorization in information retrieval introduced by Salton, Wong and Yang [3]. The goal of the VSM is to convert these textual documents to feature vectors [4].

D is a given set of documents and each document is a set of words.

Document i with vector $d_i$ can be represented as

$$d_i = (w_{1,i}, w_{2,i}, .......... w_{N,i}) \quad \text{---------------------- (1)}$$

where $w_{j,i}$ represents the weight for word j that occurs in document i and N is the number of words used for vectorization. To compute wi,j, we can set it to 1 when the word j exist in document i and 0 otherwise. The number of times the word j is observed in document i can also be recorded. A more generalized scheme named term frequenct – inverse document frequency (TF-IDF) where $w_{j,i}$ is calculated as

$$w_{j,i} = tf_{j,i} \times idf_i \quad \text{-------------------- (2)}$$

Where $tf_{j,i}$ is the frequency of word j in document i. $idf_i$ is the inverse frequency of word j across all documents,

$$idf_j = \log_2 \frac{|D|}{|\{document \in D \mid j \in document\}|} \quad \text{----------------------- (3)}$$

which is the logarithm of the total number of documents divided by the number of documents that contain word j. TF-IDF assigns higher weights to words that are less frequent across documents and at the same time have higher frequencies within the document they are used. Also which are common in all documents are assigned smaller weights.

## III. Related Works

In recent past, many researchers have used the Vector Space Model in social network analysis to extract. In [5], the authors present a VSM approach for extracting and representing relations from text corpus. The proposed approach uses VSM to represent the weight value of every social object's frequency in every text. It reflects the relationships between social objects and text corpus. The authors conclude that the application of VSM can obtain deeper social relations hid in text and text corpus and increase the effect and efficiency of social network analysis of text corpus. In [6], the authors use Improved Vector Space Model (IVSM) to effectively mine social networks of person entities from Wikipedia where a person entity was represented as a vector by anchor text set and content text set of his page in Wikipedia. In [7], the authors propose a novel method called WR-KMeans based on Extended Vector Space Model which outperforms the traditional k-means and bisecting k-means algorithms.

## IV. Proposed Models and Experimental Results

The OSN sites reflect user profiles and the ties among them. The users profile contains information about the users and their activities. In this paper, primarily we focus on the data generated through the activities of a friend or acquaintance on others contents like photographs, posts, updates etc. Our aim is to explore and analyze the association between the objects like photographs, posts etc and its viewers like friends, acquaintances etc for a given user and to find activity relationship among them. In this paper we try to capture the activities of the actors like friends, acquaintances etc. based on their actions like viewing, rating, making comments etc. performed on various posted objects like updates, photographs, documents, videos etc. posted by a user. As these data are unstructured in nature, we use the TF-IDF weighting scheme of Vector Space Model for vectorization purpose to get tabular data. After vectorization the vector data are being presented in a form of a weighted graph through Gephi software. As we see that TF-IDF assigns higher weights to words that are less frequent across documents and at the same time have higher frequencies within the document they are used. Also which are common in all documents are assigned smaller weights. Likewise the common actors across all the posted objects are assigned smaller weights and the actors who are less frequent across all the posted objects but performed multiple actions (multiple posts, likings, ratings etc) on specific object(s) are assigned higher weights.

The document strings are like:
$d_1$= {UID1, UID2, UID3}
$d_2$= {UID3, UID2, UID4}
$d_3$= {UID5, UID6, UID4}

Where $d_1, d_2, d_3$ etc are various objected posted by the users and UID1, UID2 etc are the actors.
Table – 1 shows Hence the *tf* values.

| object | UID1 | UID2 | UID3 | UID4 | UID5 | UID6 |
|---|---|---|---|---|---|---|
| $d_1$ | 1 | 1 | 1 | 0 | 0 | 0 |
| $d_2$ | 0 | 1 | 1 | 1 | 0 | 0 |
| $d_3$ | 0 | 0 | 0 | 1 | 1 | 1 |

**Table - 1**

The *idf* values are:
$idf_{UID1} = \log_2 (3/1)$ =1.584
$idf_{UID2} = \log_2 (3/2)$ =0.584
$idf_{UID3} = \log_2 (3/2)$ =0.584
$idf_{UID4} = \log_2 (3/2)$ =0.584
$idf_{UID5} = \log_2 (3/1)$ =1.584
$idf_{UID6} = \log_2 (3/1)$ =1.584

Now the TF-IDF values are calculated by multiplying the *tf* values with the the *idf* values calculated above.

| object | UID1 | UID2 | UID3 | UID4 | UID5 | UID6 |
|---|---|---|---|---|---|---|
| $d_1$ | 1.584 | 0.584 | 0.584 | 0 | 0 | 0 |
| $d_2$ | 0 | 0.584 | 0.584 | 0.584 | 0 | 0 |
| $d_3$ | 0 | 0 | 0 | 0.584 | 1.584 | 1.584 |

**Table- 2**

Table - 2 represents vector data available for various data mining task and visualization. Here UID1 has higher weight for $d_1$ as UID5 and UID6 have for $d_3$. UID3 and UID4 are common for $d_1$ and $d_2$. Higher weight denotes uncommon actor and stronger association.

Picture-1 shows the graphical representation of Table-2 using Gephi software where the bold lines indicate higher weights. Gephi produces the following statistics based on the graph:

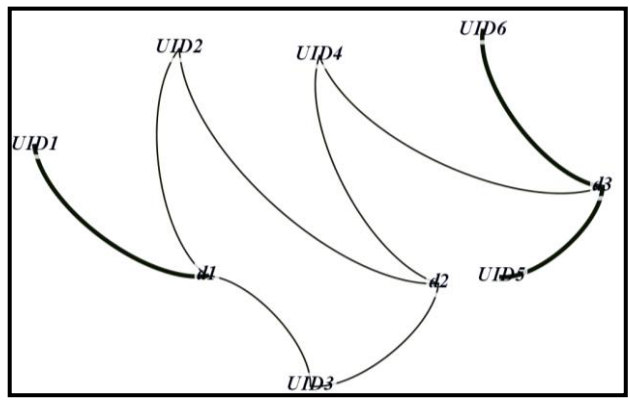

**Picture - 1**

Diameter: 6
Radius: 3
Average Path length: 2.7222
Number of shortest paths: 72
Average Weighted Degree: 1.835
Graph Density: 0.250

## V. Conclusion and Future Scope

In this paper, the TF-IDF weighting scheme of Vector Space Model has been used to analyze user activities and to find association between the objects (like photographs, documents, posts etc) and its viewers (friends, acquaintances etc) for a given user. The unstructured data of user activities has been converted and presented in tabular form. Table-2 shows the TF-IDF values against the activities performed by the various actors on different objects. The tabular data may be processed using various data mining algorithms to further analyze and visualize which may help to find better relationship and association among them.